# TOPP-DWR: Time-Optimal Path Parameterization of Differential-Driven Wheeled Robots Considering Piecewise-Constant Angular Velocity Constraints

Yong Li[1*], *Member*, *IEEE*, Yujun Huang[1], Yi Chen[1] and Hui Cheng[2], *Member*, *IEEE*

Github: https://github.com/liyong2019/TOPP-DWR.git

*Abstract*— Differential-driven wheeled robots (DWR) represent the quintessential type of mobile robots and find extensive applications across the robotic field. Most high-performance control approaches for DWR explicitly utilize the linear and angular velocities of the trajectory as control references. However, existing research on time-optimal path parameterization (TOPP) for mobile robots usually neglects the angular velocity and joint velocity constraints, which can result in degraded control performance in practical applications. In this article, a systematic and practical TOPP algorithm named TOPP-DWR is proposed for DWR and other mobile robots. First, the non-uniform B-spline is adopted to represent the initial trajectory in the task space. Second, the piecewise-constant angular velocity, as well as joint velocity, linear velocity, and linear acceleration constraints, are incorporated into the TOPP problem. During the construction of the optimization problem, the aforementioned constraints are uniformly represented as linear velocity constraints. To boost the numerical computational efficiency, we introduce a slack variable to reformulate the problem into second-order-cone programming (SOCP). Subsequently, comparative experiments are conducted to validate the superiority of the proposed method. Quantitative performance indexes show that TOPP-DWR achieves TOPP while adhering to all constraints. Finally, field autonomous navigation experiments are carried out to validate the practicability of TOPP-DWR in real-world applications.

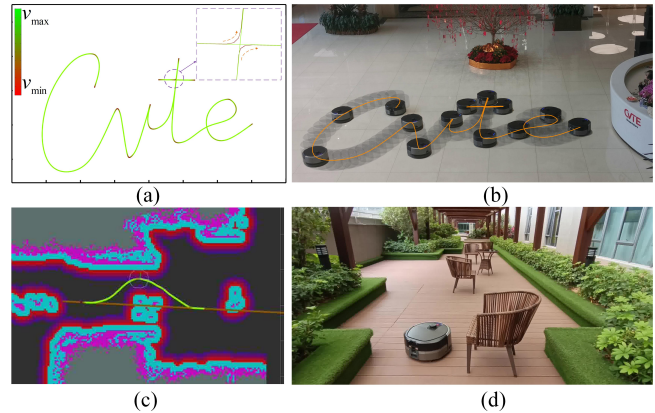

Fig. 1. (a) TOPP-DWR's time-optimal trajectory for the complex curve "cvte" in Experiment B-I; (b) Trajectory tracking results in Experiment C-I; (c) Online obstacle-avoiding trajectory in Experiment C-II; (d) The experimental scenario in Experiment C-II: a semi-outdoor garden with randomly placed chairs and table.

## I. INTRODUCTION

### A. Motivation

Mobile robots have found extensive applications in commercial and industrial fields, including transportation, cleaning, and inspection [1], [2]. Time-optimal autonomous navigation is of utmost significance for maximizing productivity [3], [4]. However, several challenges remain to be addressed, and one of them is time-optimal path parameterization (TOPP).

TOPP is defined as determining the fastest way to traverse a predefined path while satisfying system constraints [4]. Given its theoretical and practical importance, TOPP has attracted extensive attention in the research community [5], [6], [7].

The resulting time-optimal trajectory of TOPP serves as the tracking reference for motion control. To ensure satisfactory control performance, the trajectory planning algorithm should satisfy the set-out demands from motion control. Typical motion control methods for mobile robots include linear quadratic regulator (LQR) [8], [9], model predictive control (MPC) [10], [11], learning-based control [12], [13], and nolinear feedback linearization control [14], [15]. Most of these control methods explicitly utilize the linear velocity and angular velocity of the trajectory as control references [9]-[15]. However, current research on TOPP for mobile robots usually only considers longitudinal constraints, i.e., linear velocity, linear acceleration, and linear jerk constraints, and neglects angular velocity constraints, which can result in degraded control performance in practical applications.

Achieving TOPP for differential-driven wheeled robots (DWR) while adhering to linear velocity, linear acceleration, angular velocity, and joint velocity constraints is critical for maximizing the robot's working efficiency and is the primary focus of this article. Although the proposed method is developed and demonstrated with DWR, it can be readily applied to other types of mobile robots.

### B. Related Work

Approaches to TOPP can be classified into three categories: numerical integration [7], [16], dynamic programming [17], [18], and convex optimization [4], [19]. Detailed analyses and comparisons can be seen in [4], [20].

In recent years, TOPP for *n*-degree-of-freedom (*n*-DOF) robot manipulators has attracted extensive attention in the research field [20], [21]. Since the dynamic equations for the *n*-DOF robot manipulators are usually expressed in the joint space, and there is a static mapping relation between the task-space path and the joint-space one, path parameterization is usually conducted in the joint space [22], [23]. H. Pham et al. propose an approach based on reachability analysis, called TOPP-RA, in which reachable and controllable sets are recursively computed by solving small linear programming (LP) problems [4]. The asymptotic optimality of TOPP-RA is proved by I. Spasojevic et al. [24]. G. Csorvási et al. approximate the TOPP problem by sequentially solving two second-order-cone programming (SOCP) problems, and near-time-optimal planning performance can be obtained with improved computational efficiency [25]. Furthermore, L. Consolini et al. introduce virtual functions $\lambda_j$ and $\eta_j$ to transform the original system into a simple form, where the linear velocity constraints are represented as monotonic increasing functions [20]. Then, the convex problem can be solved sequentially with the optimal time complexity. However, the discrepancy from the actual system may restrict its practical applications. Research considering high-order constraints can also be found in the literature [5], [6], [22].

[1]Commercial Cleaning Robot Business Unit, Guangzhou Shiyuan Electronic Technology Co. Ltd., Guangzhou 510300, China.
[2]School of Data and Computer Science, Sun Yat-sen University, Guangzhou 510006, China.
*Corresponding author: Yong Li (e-mail: liyong2018@zju.edu.cn).

Regarding the research on TOPP for mobile robots, Z. Fan et al. extend TOPP-RA to realize cartesian-based trajectory optimization while considering joint jerk constraints [26]. Sequential convex programming is applied to TOPP for an autonomous car [27]. Longitudinal velocity and acceleration constraints are considered, and the problem is solved using a gradient-based Gauss-Newton method. Y. Zhang et al. reformulate speed planning for autonomous driving as a SOCP problem [28], considering time efficiency, smoothness objectives and dynamics, friction circle, and boundary condition constraints. During the construction of SOCP, the introduction of slack variables may double the problem size and make the approach inefficient [29]. To avoid this drawback, G. Tang et al. directly solve the nonlinear optimization problem with a primal-dual interior point method [29]. L. Consolini et al. propose an optimal complexity algorithm for an autonomous vehicle. Its optimality is theoretically guaranteed under the linear velocity, linear acceleration, and normal acceleration constraints [30].

A practical TOPP algorithm for DWR and other types of mobile robots should consider angular velocity, joint-space velocity, linear velocity and linear acceleration constraints. To the best of the authors' knowledge, no systematic approach has been found in the literature.

*C. Contributions*

This article proposes a systematic and practical TOPP algorithm for DWR. The main contributions are as follows:

(1) For the first time, a TOPP algorithm considering linear acceleration, linear velocity, angular velocity, and joint velocity constraints is proposed for DWR in the task space, which can be readily applied to other types of mobile robots.

(2) Based on the traverse time between two consecutive path points, piecewise-constant angular velocity and joint velocity constraints are converted into linear velocity constraints. A slack variable is introduced, and a SOCP problem is reformulated to solve the TOPP efficiently.

(3) Comparative experimental results and quantitative performance indexes indicate that the proposed method can achieve time-optimal trajectory planning performance for standard profile, complex curve, and full-coverage path while satisfying all constraints.

(4) Although the proposed algorithm considers all the constraints of DWR, it still has satisfactory computational efficiency. It can run on low-cost and computational-ability-constrained commercial robots. Field navigation experimental results have verified the effectiveness and practicability of the proposed method.

The rest of the article is organized as follows: Sec. II is about the problem description, followed by Sec. III presenting the methodology. Then comparative and field experimental results are described in Sec. IV. Conclusions are drawn in Sec. V.

## II. PROBLEM DESCRIPTION

*A. Kinematic Model of the Differential-Driven Wheeled Robot*

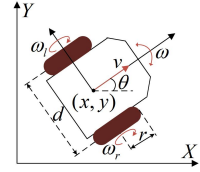

Fig. 2. Structure of the differential-driven wheeled robot.

As shown in Fig. 2, the kinematic model of the differential-driven wheeled robot describes the linear relationship between the differential of the robot's pose and the robot's velocity [1]:

$$\dot{\boldsymbol{p}} = \boldsymbol{J}_p \boldsymbol{v} \qquad (1)$$

where $\boldsymbol{p} = [x, y, \theta]^T \in R^{3\times 1}$ is the robot's pose, and $\boldsymbol{J}_p = [\cos\theta, 0; \sin\theta, 0; 0,1] \in R^{3\times 2}$ is the velocity Jacobian matrix, and $\boldsymbol{v} = [v, \omega]^T \in R^{2\times 1}$ is the robot's velocity, which is composed of the linear velocity $v$ and angular velocity $\omega$. $\boldsymbol{p}$ can be obtained by integrating $\boldsymbol{v}$:

$$\boldsymbol{p} = \boldsymbol{p_0} + \int_0^t \boldsymbol{J}_p \boldsymbol{v} dt \qquad (2)$$

where $\boldsymbol{p_0} = [x_0, y_0, \theta_0]^T \in R^{3\times 1}$ is the initial pose of the robot. The relationship between the robot's velocity and the wheel velocity $\boldsymbol{\omega}_q$ is:

$$\boldsymbol{v} = \boldsymbol{J}_\omega \boldsymbol{\omega}_q \qquad (3)$$

where $\boldsymbol{J}_\omega = [r/2, r/2; r/d, -r/d] \in R^{2\times 2}$, $\boldsymbol{\omega}_q = [\omega_r, \omega_l]^T \in R^{2\times 1}$. When considering the wheel velocity limits in the joint space, that is $\omega_r \in [\omega_r^{min}, \omega_r^{max}]$, and $\omega_l \in [\omega_l^{min}, \omega_l^{max}]$, for a given linear velocity $v_{set}$, the corresponding angular velocity $\omega(v_{set})$ is bounded with $\omega_{max}^{v_{set}}$ and $\omega_{min}^{v_{set}}$:

$$\omega_{max}^{v_{set}} = \underset{\substack{\omega_r \in [\omega_r^{min}, \omega_r^{max}] \\ \omega_l \in [\omega_l^{min}, \omega_l^{max}]}}{\operatorname{argmax}} \{\omega: \boldsymbol{v} = \boldsymbol{J}_\omega \boldsymbol{\omega}_q, v = v_{set}\} \qquad (4)$$

$$\omega_{min}^{v_{set}} = \underset{\substack{\omega_r \in [\omega_r^{min}, \omega_r^{max}] \\ \omega_l \in [\omega_l^{min}, \omega_l^{max}]}}{\operatorname{argmin}} \{\omega: \boldsymbol{v} = \boldsymbol{J}_\omega \boldsymbol{\omega}_q, v = v_{set}\} \qquad (5)$$

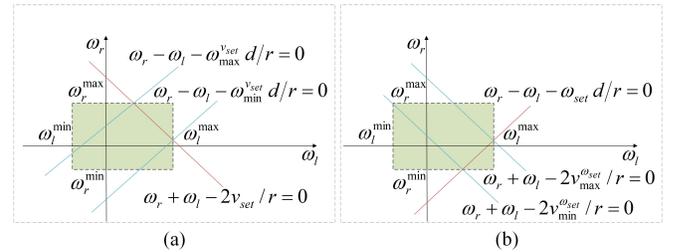

Fig. 3. When considering wheel velocity limits in the joint space: (a) for the given linear velocity $v_{set}$, the maximum and minimum feasible angular velocities are $\omega_{max}^{v_{set}}$ and $\omega_{min}^{v_{set}}$, respectively; (b) for the given angular velocity $\omega_{set}$, the maximum and minimum feasible linear velocities are $v_{max}^{\omega_{set}}$ and $v_{min}^{\omega_{set}}$, respectively.

Similarly, for a given angular velocity $\omega_{set}$, the corresponding linear velocity $v(\omega_{set})$ is bounded with $v_{max}^{\omega_{set}}$ and $v_{min}^{\omega_{set}}$:

$$v_{max}^{\omega_{set}} = \underset{\substack{\omega_r \in [\omega_r^{min}, \omega_r^{max}] \\ \omega_l \in [\omega_l^{min}, \omega_l^{max}]}}{\mathrm{argmax}} \{v: \boldsymbol{v} = \boldsymbol{J_\omega \omega_q}, \omega = \omega_{set}\} \quad (6)$$

$$v_{min}^{\omega_{set}} = \underset{\substack{\omega_r \in [\omega_r^{min}, \omega_r^{max}] \\ \omega_l \in [\omega_l^{min}, \omega_l^{max}]}}{\mathrm{argmin}} \{v: \boldsymbol{v} = \boldsymbol{J_\omega \omega_q}, \omega = \omega_{set}\} \quad (7)$$

Equations (4)-(7) describe the linear relationships between the linear velocity $v$ and the angular velocity $\omega$ when considering the wheel velocity limits in the joint space, which can be solved by constructing a linear programming problem, or by using the phase-plane shown in Fig. 3. For a given linear velocity $v_{set}$, the relationship between it and $\omega_r, \omega_l$ is represented by the straight line $\omega_r + \omega_l - 2v_{set}/r = 0$ in Fig. 3(a). $\omega_{max}^{v_{set}}$ and $\omega_{min}^{v_{set}}$ are the $\omega$ values when the straight line $\omega_r - \omega_l - \omega d/r = 0$ intersects the $\omega_r$ axis at the maximum and minimum intersection distances, respectively. Similarly, for a given angular velocity $\omega_{set}$, the relationship between it and $\omega_r, \omega_l$ is represented by the straight line $\omega_r - \omega_l - \omega_{set} d/r = 0$ in Fig. 3(b). $v_{max}^{\omega_{set}}$ and $v_{min}^{\omega_{set}}$ are the linear velocities when the straight line $\omega_r + \omega_l - 2v/r = 0$ intersects the $\omega_r$ axis at the maximum and minimum intersection distances, respectively.

*B. Path Parameterization in the Task Space*

Different from the *n*-DOF robot manipulator systems, the differential-driven mobile robot is a typical under-actuated system without static mapping between the task-space path and the joint-space one. According to (2) and (3), the pose in the task space can only be obtained by integrating joint-space velocity. To avoid the path error caused by velocity integration, path parameterization is carried out directly in the task space in this article.

## III. METHODOLOGY

*A. Path Description based on Non-Uniform Rational B-spline*

The resulting path from the path-planning is usually a set of discrete points $\mathfrak{P} = \{\mathfrak{p}_0, \mathfrak{p}_1, \ldots, \mathfrak{p}_{\ell-1}\} \in R^{\ell \times 1}$ [2], [31]. The typical approach in the literature is to use smooth curves to fit $\mathfrak{P}$ to obtain a dense and feasible initial trajectory for the subsequent path parameterization [4], [19]. In this article, the non-uniform rational B-spline curve is introduced.

The 2D $\wp$-degree B-spline curve for fitting $\mathfrak{P}$ is defined as follows [32]:

$$\boldsymbol{C}(u) = \sum_{i=0}^{\ell+\wp-2} \boldsymbol{N}_{i,\wp}(u)\boldsymbol{Q}_i = [p_x, p_y]^T \quad (8)$$

where $\boldsymbol{Q} = [Q_0, Q_1, \ldots, Q_{\ell+\wp-2}]^T \in R^{(\ell+\wp-1) \times 2}$ is the control points set. $\boldsymbol{U} = \{u_0, u_1, \ldots, u_m\}^T \in R^{(m+1) \times 1}$ is the non-decreasing knot vector with $u$ being the knot and $m = \ell + 2\wp - 1$. $\boldsymbol{N}_{i,\wp}(u)$ is defined as the $i$-th $\wp$-degree B-spline basis function at knot $u$, which can be solved recursively:

$$\boldsymbol{N}_{i,\wp}(u) = \frac{u - u_i}{u_{i+\wp} - u_i} \boldsymbol{N}_{i,\wp-1}(u) + \frac{u_{i+\wp+1} - u}{u_{i+\wp+1} - u_{i+1}} \boldsymbol{N}_{i+1,\wp-1}(u) \quad (9)$$

$$\text{with } \boldsymbol{N}_{i,0}(u) = \begin{cases} 1, u_i \leq u \leq u_j \\ 0, \text{ otherwise} \end{cases} \quad (10)$$

The tangent vector $\boldsymbol{C}'(u)$ at knot $u$ is defined as:

$$\boldsymbol{C}'(u) = \sum_{i=0}^{\ell+\wp-2} \boldsymbol{N}'_{i,\wp}(u)\boldsymbol{Q}_i$$
$$= \sum_{i=0}^{\ell} \wp \left( \frac{N_{i,\wp-1}(u)}{u_{i+\wp} - u_i} - \frac{N_{i+1,\wp-1}(u)}{u_{i+\wp+1} - u_{i+1}} \right) \boldsymbol{Q}_i = [t_x, t_y]^T \quad (11)$$

Then the orientation $\mathcal{R}(u)$ of the point at knot $u$ is expressed as:

$$\mathcal{R}(u) = \arctan(t_y / t_x) \quad (12)$$

Considering the complexity in real applications, the path point intervals in $\mathfrak{P}$ are usually nonlinear. To get guaranteed path-fitting performance, we use the chord-length sampling method to determine the distributions of $u_i$ in the knot vector [33]:

$$u_i = \begin{cases} (i - \wp)u_0, & i \in [0, \wp] \\ u_{i-1} + u_0 \frac{d_i}{d_0}, & i \in [\wp + 1, \ell + \wp - 1] \\ u_{\ell+1}, & i \in [\ell + \wp, m] \end{cases} \quad (13)$$

where $u_0$ represents the time taken to traverse the Euclidean distance of $d_0$, $d_i = \|\mathfrak{p}_{i-\wp} - \mathfrak{p}_{i-\wp-1}\|$ the Euclidean distance between $\mathfrak{p}_{i-\wp}$ and $\mathfrak{p}_{i-\wp-1}$.

*B. Task-Space Trajectory Initialization*

In the literature, a typical approach for trajectory parameterization is to describe the initial path $\mathcal{P}$ as a function of the intermediate variable $s$ as $\mathcal{P}(s)_{s \in [0, s_{end}]}$ [4], [7]. The process of trajectory parameterization is to find a nonlinear mapping from time $t$ to $s$: $[0, T] \to [0, s_{end}]$ so that the trajectory is represented as a function with respect to time: $\mathcal{P}(s(t))_{t \in [0, T]}$. During the process, $ds/dt$ and $d^2s/dt^2$ are solved explicitly for a second order continuous path $\mathcal{P}$.

In this article, a straightforward method is adopted. We sample along $\boldsymbol{U}_{u \in [u_{\wp+1}, u_{m-\wp-1}]}$ with a fixed resolution $\Delta_u$ to get an initial time sequence $\boldsymbol{t} = \{t_0, t_1, \ldots, t_{n-1}\} \in R^{n \times 1}$ with $t_0 = 0$ and $t_{n-1} = u_{m-\wp-1}$. Then, the initial trajectory for path parameterization can be expressed as $\mathcal{P} = \{p_0, p_1, \ldots, p_{n-1}\} \in R^{n \times 1}$, where $p_i = \{x_i, y_i, \theta_i\}^T = \{\boldsymbol{C}(t_i), \mathcal{R}(t_i)\}$. During the optimization process, the optimal time $\boldsymbol{t}$ is solved directly.

*C. Optimization Problem Description*

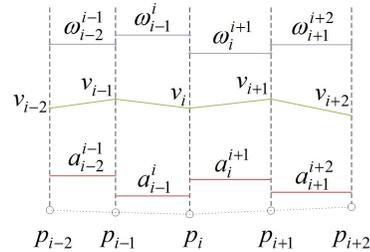

Fig. 4. The piecewise-constant linear acceleration $a$, piecewise-linear linear velocity $v$, and piecewise-constant angular velocity $\omega$ in the optimization. $a_i^{i+1}$ and $\omega_i^{i+1}$ represent the linear acceleration and angular velocity within the time interval $[i, i + 1]$, respectively.

As shown in Fig. 4, $a_i^{i+1}$ and $\omega_i^{i+1}$, short for $a_i$ and $\omega_i$ in the subsequent descriptions, represent the linear acceleration and angular velocity within the time interval $[i, i + 1]$, respectively. For the convenience of notation, piecewise-constant and piecewise-linear are shorted as $PC$ and $C^1$,

respectively. The TOPP problem in this article is based on the following assumption:

*Assumption 1:* $a$, $v$, and $\omega$ are $PC$, $C^1$, and $PC$, respectively.

Adopting the $PC$ $a$ and $C^1$ $v$ is a typical approach in time-optimal trajectory planning. However, such practice is not applied to $\omega$. Only $PC$ $\omega$ is considered in this article for the following reasons: 1) Compared with longitudinal motion, the discontinuity in $\omega$ has less impact on motion smoothness and navigation safety; 2) Introducing $C^1$ $\omega$ will increase the complexity of the algorithm and brings extra computational burden, which will restrict the application of the proposed method to low-cost and computational-ability-constricted robots.

*1) Optimization Objective.* Considering *Assumption 1* and let $t_f = t_{n-1}$ be the total time of the trajectory, the optimization objective can be described as:

$$\mathcal{J} = \min_{a, \omega \in PC[0, t_f], v \in C^1[0, t_f]} t_f \quad (14)$$

*2) Task-Space Constraints.* The following task-space constraints are considered in the optimization:

$$0 \le v \le v^{max} \quad (15)$$
$$a^{min} \le a \le a^{max} \quad (16)$$
$$\omega^{min} \le \omega \le \omega^{max} \quad (17)$$
$$0 \le a_n = v^2 \kappa \le a_n^{max} \quad (18)$$

where $[0, v^{max}]$, $[a^{min}, a^{max}]$, and $[\omega^{min}, \omega^{max}]$ are the boundaries of $v$, $a$, and $\omega$, respectively, $\kappa$ the curvature of the path. The adoption of $a_n$ helps to adjust $v$ proportionally with $\kappa$ [30]. Combing (15) and (18), the constraints of $v$ can be collectively expressed as:

$$0 \le v \le \bar{v} \quad (19)$$

where $\bar{v} = min(v^{max}, \sqrt{a_n^{max}/\kappa})$.

*3) Joint-Space Velocity Constraints.* The joint-space constraints are determined mainly by the dynamics of the joint actuators. Moreover, joint-space velocity constraints are considered in this article:

$$\underline{v_r} \le v_r \le \overline{v_r} \quad (20)$$
$$\underline{v_l} \le v_l \le \overline{v_l} \quad (21)$$

where $v_r = \omega_r r$, $v_l = \omega_l r$, $\overline{v_r} = \omega_r^{max} r$, $\underline{v_r} = \omega_r^{min} r$, $\overline{v_l} = \omega_l^{max} r$, $\underline{v_l} = \omega_l^{min} r$.

*4) Boundary Constraints.* The boundary constraints aim to meet the trajectory's linear velocity and angular velocity constraints at the initial and terminal states.

$$v_0 = v_s, v_{n-1} = v_f, \omega_0 = \omega_s, \omega_{n-1} = \omega_f \quad (22)$$

To make the optimization problem feasible, the following assumption should be made.

*Assumption 2:* $v_0$ and $v_f$ are within $[0, \bar{v}]$, while $\omega_0$ and $\omega_f$ are bounded within $[\omega^{min}, \omega^{max}]$, respectively.

*5) Optimization Problem Discretization and Reformulation.* To numerically solve (14)-(22), it is necessary to discretize and reformulate the aforementioned problem. For any two consecutive time instants $t_k$ and $t_{k+1}$, $k = 0, \dots, n-2$, the corresponding path points are $p_k$ and $p_{k+1}$, let $\Delta s_k = \|p_{k+1} - p_k\|$ and $\Delta \theta_k = \theta_{k+1} - \theta_k$. Based on *Assumption 1*, the time interval $\Delta t_k = t_{k+1} - t_k$ can be expressed as:

$$\Delta t_k = \frac{2\Delta s_k}{v_k + v_{k+1}}, k \in [0, n-2] \quad (23)$$

$$\Delta t_k = \frac{\Delta \theta_k}{\omega_k}, k \in [0, n-2] \quad (24)$$

Considering (23) and the definition of linear acceleration, $a_k$ can be expressed as:

$$a_k = \frac{v_{k+1} - v_k}{\Delta t_k} = \frac{v_{k+1}^2 - v_k^2}{2\Delta s_k}, k \in [0, n-2] \quad (25)$$

Then, the linear acceleration constraints in (16) can be represented as:

$$\underline{a_k} \le v_{k+1}^2 - v_k^2 \le \overline{a_k}, k \in [0, n-2] \quad (26)$$

where $\overline{a_k} = 2\Delta s_k a^{max}$, $\underline{a_k} = 2\Delta s_k a^{min}$. Combining (23) and (24), the relationship between $\omega_k$ and $v_k, v_{k+1}$ can be represented as:

$$\omega_k = \frac{\Delta \theta_k}{2\Delta s_k}(v_k + v_{k+1}), k \in [0, n-2] \quad (27)$$

Considering the asymmetry of the angular velocity boundaries, constraints (17) can be represented as:

$$\underline{\omega_k} \le v_k + v_{k+1} \le \overline{\omega_k}, k \in [0, n-2] \quad (28)$$

where $\underline{\omega_k} = min(h_k \omega^{max}, h_k \omega^{min})$ and $\overline{\omega_k} = max(h_k \omega^{max}, h_k \omega^{min})$, $h_k = 2\Delta s_k/\Delta \theta_k$. Substituting (3), (27) into (20) and (21), the joint-space velocity constraints can be represented as:

$$\underline{v_r} \le (1 + g_k)v_k + g_k v_{k+1} \le \overline{v_r}, k \in [0, n-2] \quad (29)$$

$$\underline{v_l} \le (1 - g_k)v_k - g_k v_{k+1} \le \overline{v_l}, k \in [0, n-2] \quad (30)$$

where $g_k = d\Delta \theta_k/(4\Delta s_k)$. Substituting (23) into (14) and the optimization objective can be discretized as the function of $v_k$:

$$\mathcal{J} = \min_{v_k} \sum_{k=0}^{n-2} \frac{2\Delta s_k}{v_k + v_{k+1}} \quad (31)$$

*6) SOCP Problem Construction.* The objective function (31) is non-convex and to solve it efficiently, a slack variable $c_k$ will be introduced, and an additional slack constraint will be added:

$$c_k \ge \frac{1}{v_k + v_{k+1}}, k \in [0, n-2] \quad (32)$$

Equation (32) can be rewritten in the form of the Rotated Quadratic Cone constraint:

$$(c_k, v_k + v_{k+1}, \sqrt{2}) \in \mathcal{Q}_r^3, k \in [0, n-2] \quad (33)$$

With the introduction of $c_k$, the optimization objective (31) can be reformulated as a convex function:

$$\mathcal{J} = \min_{v_k, c_k} \sum_{k=0}^{n-2} 2\Delta s_k c_i \quad (34)$$

To solve the problem more efficiently, the liner acceleration constraints in (26) will be reformulated as convex functions. Substituting (32) into (26), the linear acceleration constraints can be represented as:

$$v_{k+1} - v_k \le \frac{\overline{a_k}}{v_k + v_{k+1}} \le \overline{a}_k c_k, k \in [0, n-2] \quad (35)$$

$$v_k - v_{k+1} \leq -\frac{a_k}{v_k + v_{k+1}} \leq -\underline{a}_k c_k, k \in [0, n-2] \quad (36)$$

Considering the optimization objective (34) and the constraints (19), (22), (28), (29), (30), (33), (35), and (36), the TOPP can be described as the following discrete-form SOCP problem:

$$\mathcal{J} = \min_{v_k, c_k} \sum_{k=0}^{n-2} 2\Delta s_k c_i \quad (37)$$

$$\text{subject to } 0 \leq v_k \leq \bar{v}_k, k \in [0, n-1] \quad (38)$$

$$v_{k+1} - v_k \leq \bar{a}_k c_k, k \in [0, n-2] \quad (39)$$

$$v_k - v_{k+1} \leq -\underline{a}_k c_k, k \in [0, n-2] \quad (40)$$

$$\underline{\omega_k} \leq v_k + v_{k+1} \leq \overline{\omega_k}, k \in [0, n-2] \quad (41)$$

$$\underline{v_r} \leq (1 + g_k)v_k + g_k v_{k+1} \leq \overline{v_r}, k \in [0, n-2] \quad (42)$$

$$\underline{v_l} \leq (1 - g_k)v_k - g_k v_{k+1} \leq \overline{v_l}, k \in [0, n-2] \quad (43)$$

$$(c_k, v_k + v_{k+1}, \sqrt{2}) \in \mathcal{Q}_r^3, k \in [0, n-2] \quad (44)$$

$$v_0 = v_s, v_{n-1} = v_f, \omega_0 = \omega_s, \omega_{n-1} = \omega_f \quad (45)$$

*Discussion:* In light of the state-of-the-art algorithms in robotic path parameterization, if the linear velocity constraints can be expressed as $v_i \leq f_j^i(v_{i+1}), v_{i+1} \leq b_k^i(v_i)$, where $f$ and $b$ are both monotonically decreasing convex functions, then the problem can be solved in a sequential form with the optimal time complexity[20], [21], [30]. However, as seen in (41)-(43), the constraints can not be represented as the form required by sequential linear programming, which indicates the complexity and uniqueness of the problem investigated in this article.

## IV. EXPERIMENTS

### A. Experiment with Standard Lissajous Curve

To verify the effectiveness of the proposed method, the standard Lissajous profile shown in (46) and Fig. 5 is adopted firstly.

$$\begin{cases} x = 10(\cos(\pi/4) - \cos(3\pi t/50 + \pi/4)) \\ y = 2(1 - \cos(2\pi t/50)) \end{cases} \quad (46)$$

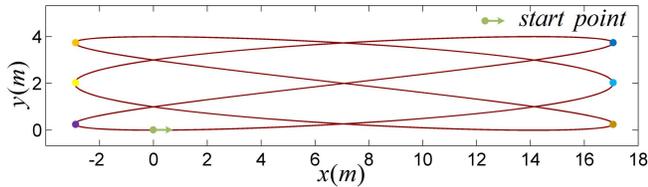

Fig. 5. The Lissajous curve in Experiment A. The six colored points feature regions with large curvatures.

Since the Lissajous curve is $C^n$, there is no need to use the B-spline defined in (8) for curve fitting. We sample within the time interval of $[0, \mathcal{t}]$, with $\mathcal{t} = 100s$ the period of the curve, at the fixed time resolution of $0.01s$ to obtain the initial trajectory $\mathcal{P}$. The corresponding linear velocity profile is shown in Fig. 6. In the experiment, $v^{max} = 0.6m/s$, $a^{max} = 1m/s^2$, $a^{min} = -1m/s^2$, $a_n^{max} = 0.6m/s^2$, $\omega^{max} = 2rad/s$, $\omega^{min} = -2rad/s$, $\overline{v_r} = \overline{v_l} = 0.75m/s$, $\underline{v_r} = \underline{v_l} = -0.75m/s$, $d = 0.35m$, $v_0 = v_f = 0m/s$, $\omega_0 = \omega_f = 0rad/s$. The computation unit is an industrial computer with the CPU *i7-10700* @*2.9GHz×16* and the RAM of 32GB. The C language API of *Mosek 10.2* is introduced to solve the optimization problem.

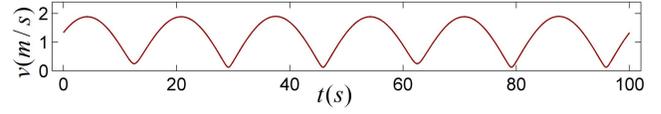

Fig. 6. The initial linear velocity profile in Experiment A.

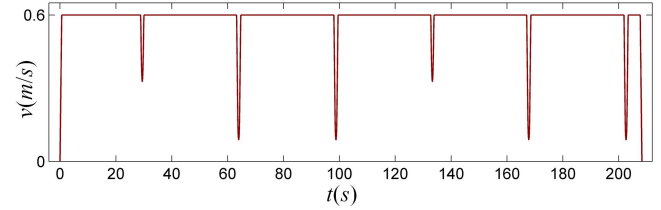

Fig. 7. The resulting linear velocity profile in Experiment A.

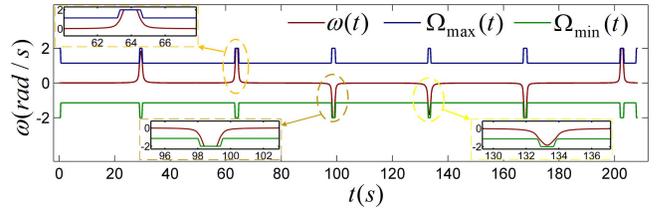

Fig. 8. The resulting angular velocity profile in Experiment A. $\Omega_{\max}(t)$ and $\Omega_{\min}(t)$ are the upper and lower bounds of $\omega(t)$, respectively. The curves in the orange, chocolate, and yellow enlarged windows correspond to the orange, chocolate, and yellow points regions in Fig. 5, respectively.

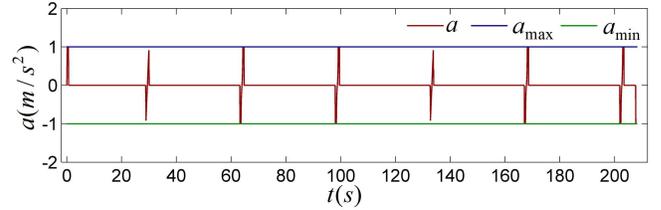

Fig. 9. The resulting linear acceleration profile in Experiment A.

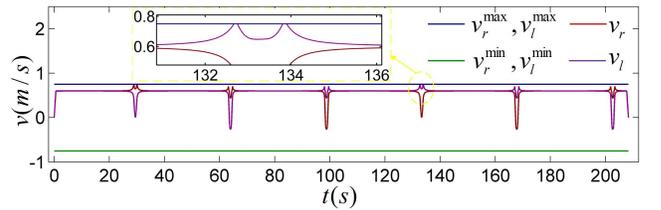

Fig. 10. The resulting joint velocity profile in Experiment A. The curves in the yellow enlarged window correspond to the yellow point region in Fig. 5.

The maximum linear velocity of the initial trajectory is $1.894m/s$, exhibiting a significant discrepancy from $v^{max} = 0.6m/s$. Nevertheless, the desired linear velocity profile is successfully acquired as illustrated in Fig. 7. In Fig. 7, excluding the initial and terminal states, there are six prominent low-velocity regions, which closely correspond to the six colored points in Fig.5. Larger curvatures (the orange, chocolate, blue and dark violet points in Fig.5) result in lower velocity, which aligns with the constraint in (38).

For any $t_k$ and the corresponding $v_i$ in Fig. 7, the boundary angular velocity $\Omega_{\max}$ and $\Omega_{\min}$ for $\omega_i$ in Fig. 8 can be ob-tained by combining (4), (5), and (17): $\Omega_{max} = min(\omega_{max}^{v_i}, \omega_{max})$, $\Omega_{min} = max(\omega_{min}^{v_i}, \omega_{min})$. As seen from Fig. 5 and 8, due to the largest curvatures in the orange and chocolate enlarged windows, $\omega_i$ reach $\Omega_{max}$ and $\Omega_{min}$, respectively. This article explicitly describes linear acceleration constraints as convex constraints by introducing the slack variable $c_k$. As shown in Fig. 9, the resulting $a$ is consistent remains within the boundary.

An interesting phenomenon is observed in Fig. 9 and the joint-space velocity profile in Fig.10. The linear accelerations of the four regions with the largest curvatures in Fig.5 reach the linear acceleration boundary. In contrast, the two with minor curvatures reach the joint velocity boundaries. This phenomenon indicates that for the optimal-time trajectory, it is not always the case that the larger the curvature, the faster the joint velocity.

Furthermore, as depicted in Fig. 10, the joint velocity profiles are more closely approaching the upper boundary than the lower one. This phenomenon is explainable. As illustrated in Fig. 3(b), when $\omega = -2rad/s$, $v = 0m/s$, the minimum feasible $\omega_r$ is derived at the intersection point of the lines $\omega_r - \omega_l - \omega d/r = 0$ and $\omega_r + \omega_l - 2v/r = 0$, with the resulting minimum $v_r = \omega_r r$ being $-0.35 m/s$; Similarly, when $\omega = 2rad/s$ and $v = 0m/s$, the minimum $v_l = -0.35 m/s$ is obtained.

*B. Comparative Experiments*

To the authors' best knowledge, in the existing literature, there has been scant practical and systematic research on path parameterization methods for mobile robots considering linear velocity, angular velocity, and joint-space velocity constraints in the task space. Nevertheless, to demonstrate the superiority of the proposed algorithm, we compare our method with the most relevant algorithms in the literature.

*TOPP-DWR*: The algorithm proposed in this article. The parameters and hardware platform employed in Experiment A are also utilized in this experiment. The sole difference is that the initial path with an approximate interval of $0.2m$ will be curve-fitted by the B-spline curve (8) to function as the initial trajectory, where the sampling interval is $\Delta_t = 0.05s$. To make the comparison fair, the subsequent algorithms share the same parameters as those in TOPP-DWR.

*SLP*: The sequential linear programming algorithm proposed by L. Consolini et al. [20], [30]. If linear velocity constraints can be represented as monotonically decreasing convex functions, the problem can be solved sequentially with linear time complexity. However, SLP for mobile robots [30] does not consider angular or joint velocity constraints.

*BKA*: The B-spline knot allocation algorithm proposed by B. Zhou et al. [34]. If the linear velocity, linear acceleration, and angular velocity calculated based on (8) exceed the predefined thresholds, the knots of the B-spline curve will be reallocated proportionally.

*OTAQ*: The time-optimal allocation algorithm for quadrotor proposed by F. Gao et al. [19]. The temporal trajectory is represented by piecewise polynomials. A SOCP problem is constructed with considering linear velocity and acceleration constraints. The experiment is carried out based on the open-source code provided by the authors. In the experiment, $\rho = 0$. $\Delta_s = 0.2$ is selected so that the numbers of final trajectory points of OTAQ and TOPP-DWR are similar for the same initial path. To facilitate comparison, the linear velocity boundary is adjusted to be the boundary for $\|v\|$ as in (15) rather than for $v_x$, $v_y$, and $v_z$, respectively.

BKA and OTAQ are designed for quadrotors and joint space velocity constraints are not considered. Therefore, the path parameterization performance in $a$, $v$, and $\omega$ are compared in particular. To quantify the performance of different algorithms, the following performance indexes are introduced: 1) the computation time $t_c$; 2) the trajectory time $t_f$, which is used to evaluate the optimality of the trajectory; 3) The normalized maximum linear velocity $\varsigma = max(v_k/\bar{v}_k)$; 4) the normalized maximum linear acceleration $\varrho = max(a_i/a^{max}, a_i/a^{min})$; 5) the normalized maximum angular velocity $\chi = max(\omega_i/\Omega_{max}, \omega_i/\Omega_{min})$. If any of the values of $\varsigma$, $\varrho$, and $\chi$ exceed 1, the constraints are violated [7].

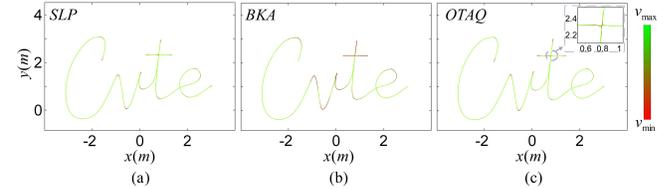

Fig. 11. The resulting trajectories in Experiment B-I.

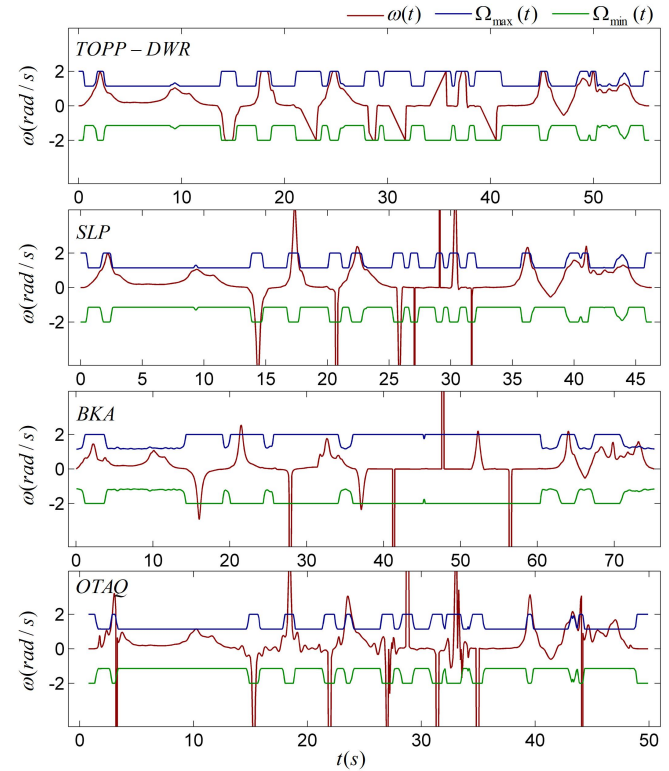

Fig. 12. The resulting angular velocity profiles in Experiment B-I.

*Experiment I:* The target curve in the experiment is shown in Fig. 11 with the complex pattern of "cvte." The path length is $24.389m$. The angular velocity profile and performance indexes are shown in Fig. 12 and Table I, respectively. As seen from Fig. 1(a), Fig. 11, and Table I, TOPP-DWR and SLP exhibit comparable linear velocity and acceleration performance. The primary distinction is that SLP does not consider angular velocity or joint-space velocity constraints. Within the trajectory time of $46.482s$, SLP has as many as 11 evident angular-velocity-boundary violations, with the maximum angular velocity reaching 126.531$rad/s$. The knot re-allocation strategy for BKA is simple and fast but has the poorest performance indexes, as shown in Table I. As shown in the enlarged window in Fig. 11(c), OTAQ presents a phenomenon of a tortuous fitting result, which may be related to the excessive complexity of the curve. Since angular velocity constraints are not considered, the resulting angular velocity is far from satisfactory.

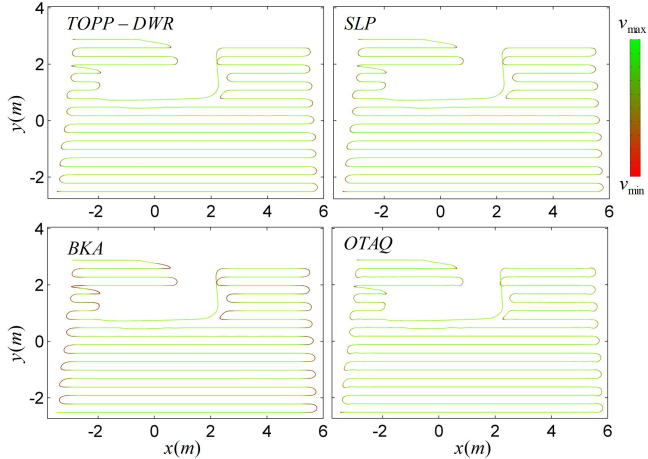

Fig. 13. The resulting trajectories in Experiment B-II.

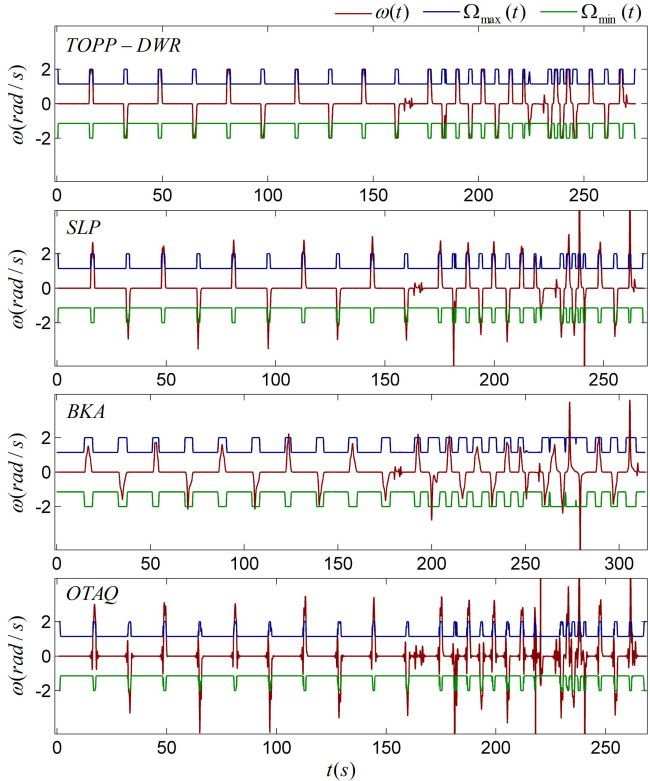

Fig. 14. The resulting angular velocity profiles in Experiment B-II.

*Experiment II:* To verify the path parameterization performance for paths that are from real-world applications, Exp. II is conducted based on a typical full coverage path of a commercial cleaning robot named "C3". The comparison results of linear velocities, angular velocities, and performance indexes are shown in Fig. 13, 14, and Table I, respectively. The results of Exp. II are basically consistent with those of Exp. I. Compared with SLP and BKA, TOPP-DWR demands more computational resources to achieve TOPP performance considering linear, angular, and joint constraints. With the path length of 150.531$m$, the number of path points in the final trajectories of OTAQ and TOPP-DWR is 3820 and 5094, respectively. For the total planning time of 102.430$s$ for OTAQ, the piecewise polynomial curve fitting and path parameterization take 101.822$s$ and 0.608$s$, respectively. In contrast, for the total planning time of 0.447$s$ for TOPP-DWR, B-spline curve-fitting takes 0.105$s$, and path parameterization 0.342$s$. Even though TOPP-DWR considers additional angular velocity and joint velocity constraints, it takes 43.75% less optimization time than OTAQ. One possible reason is that OTAQ's optimization process introduces extra continuous constraints of linear velocity and acceleration for piecewise polynomials.

TABLE I. PERFORMANCE INDEXES IN EXPERIMENT B I-II

| Exp. | | $t_c$ | $t_f$ | $\varsigma$ | $\varrho$ | $\chi$ |
|---|---|---|---|---|---|---|
| **B-I** | Ours* | 0.070 | 54.901 | **1.0** | **1.0** | **1.0** |
| | SLP | **0.0343** | **46.482** | **1.0** | **1.0** | 126.531 |
| | BKA | 0.0593 | 72.672 | 0.999 | 1.016 | 31.416 |
| | OTAQ | 0.628 | 48.443 | **1.0** | 1.243 | 18.204 |
| **B-II** | Ours* | 0.447 | 274.428 | **1.0** | **1.0** | **1.0** |
| | SLP | 0.275 | **268.418** | **1.0** | **1.0** | 7.665 |
| | BKA | **0.216** | 313.883 | 1.003 | 0.958 | 3.573 |
| | OTAQ | 102.430 | 267.305 | **1.0** | 1.0864 | 81.909 |

*Ours*: The TOPP-DWR algorithm proposed in this article

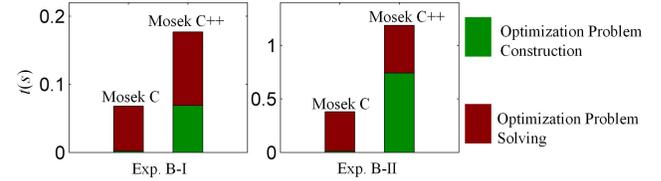

Fig. 15. Analysis of the computation time composition in the optimization stage for Mosek C and Mosek C++ in Experiments B-I and B-II.

It should be noted that in the comparative experiments, we employ the C language API of the commercial solver Mosek 10.2 to solve the SOCP optimization problem. Furthermore, we also evaluated the performance of Mosek 10.2's C++ API. With the optimization results remaining consistently identical, the computation time analysis presented in Fig. 15 indicates that Mosek's C++ API exhibits superior computational efficiency. This is primarily attributed to the negligible time spent on the optimization problem construction process.

*C. Field Navigation Experiment*

To validate the applicability of TOPP-DWR to low-cost and computational-ability-constrained robots, field autonomous navigation experiments are carried out. The commercial cleaning robot used in the experiments is manufactured by CVTE, whose computation unit is based on RK3399. During the experiments, trajectories generated by TOPP-DWR serve as the tracking references for a nonlinear controller based on feedback linearization.

Field autonomous navigation experiment I is performed in a hall, with the results in Experiment B-I serving as the global reference trajectory. The computation time for TOPP-DWR on RK3399 is 0.726s. Tracking results are shown in Fig. 1 (b), with the root-mean-square tracking error of 0.0156$m$.

To further evaluate the applicability of TOPP-DWR in dynamic environments, field navigation experiment II is carried out in a semi-outdoor garden, as shown in Fig. 1(d). After post-processing [31], the global path between the start and the goal is a straight line. As shown in Fig. 1(c), several thin-legged chairs are randomly placed during the experiment to block the way. The average computation time for TOPP-DWR in online obstacle-avoiding is $0.0728s$, which can meet the set-out demands in practical applications. Videos for the field navigation experiments are provided as supplementary materials.

## V. Conclusion

This article proposes a comprehensive and systematic time-optimal path parameterization algorithm for differential-driven wheeled robots. Distinct from the prevalent approaches in the literature, piecewise-constant angular velocity and joint velocity constraints are incorporated into the optimization problem. Moreover, the optimization problem is constructed directly, eschewing the introduction of the intermediate variable between the path and time. A SOCP problem is constructed to achieve guaranteed computational efficiency. The superiority and practicality of the proposed method is validated through comparative and field navigation experiments.

In the future, the feasibility of the optimization problem will be proved and explored. To satisfy the requirements emerging from controller design, more complex constraints, such as piecewise-linear angular velocity, will be taken into account. Our ultimate objective is to realize the time-optimal high-precision autonomous navigation for low-cost commercial robots at their physical limits in real-world applications.